\documentclass{article}

\usepackage{arxiv}

\usepackage[utf8]{inputenc} 
\usepackage[T1]{fontenc}    
\usepackage{hyperref}       
\usepackage{url}            
\usepackage{booktabs}       
\usepackage{amsmath,amsfonts,bm} 
\usepackage{nicefrac}       
\usepackage{microtype}      
\usepackage{fancyhdr}       
\usepackage{graphicx}       
\graphicspath{{figs/}}      

\usepackage{natbib}
\usepackage{tabularx}
\usepackage[flushleft]{threeparttable}
\usepackage{tcolorbox}
\usepackage{algorithm}
\usepackage{algorithmic}
\usepackage{float}
\usepackage{multirow}
\usepackage{array}

\newcommand{\RN}[1]{%
  \textup{\uppercase\expandafter{\romannumeral#1}}%
}

\title{Geo-ORBIT: A Federated Digital Twin Framework for Scene-Adaptive Lane Geometry Detection}

\author{
  Rei Tamaru \\
  Department of Civil and Environmental Engineering \\
  University of Wisconsin--Madison \\
  Madison, Wisconsin 53706, USA \\
  \texttt{tamaru@wisc.edu} \\
  \And
  Pei Li \\
  Department of Civil and Architectural Engineering and Construction Management \\
  University of Wyoming \\
  Laramie, Wyoming 82071, USA \\
  \texttt{pei.li@uwyo.edu} \\
  \And
  Bin Ran \\
  Department of Civil and Environmental Engineering \\
  University of Wisconsin--Madison \\
  Madison, Wisconsin 53706, USA \\
  \texttt{bran@wisc.edu} \\
}

\begin{document}
\maketitle

\begin{abstract}
Transportation digital twins (DTs) can support proactive traffic management and operations by maintaining dynamic virtual representations of transportation systems. Developing transportation DTs requires accurate lane-level geometry, which is commonly obtained from static maps or dedicated sensors. However, these sources limit scalability and may fail to reflect changes in roadway configuration in a timely manner. Moreover, transportation DTs rely on large volumes of multi-source data, and centralizing these data across locations raises privacy and communication concerns. To address these limitations, we introduce Geo-ORBIT (Geometrical Operational Roadway Blueprint with Integrated Twin), a decentralized, privacy-aware framework that adapts to geometry changes. Geo-ORBIT combines label-free lane detection from roadside cameras, DT synchronization, and federated meta-learning. Its core model, GeoLane, infers lane geometry from vehicle trajectories without human labels. Meta-GeoLane adapts the detection parameters to each scene, while FedMeta-GeoLane learns a shared adaptation across sites under federated updates that keep raw data local. Geo-ORBIT connects to SUMO and CARLA to render highway scenes and replay traffic, and supports a closed-loop update in which detected changes in lane geometry are propagated to the running simulation. Validated against independent human annotations on real-world data, the detected geometry achieves an error of approximately 3 m, representing nearly an order-of-magnitude improvement over the public map that supervises training. Moreover, FedMeta-GeoLane transfers a single global model across sites without per-site training and reduces per-camera communication by more than 99\% relative to centralizing trajectories. The framework is publicly available at \url{https://github.com/raynbowy23/FedMeta-GeoLane.git}.
\end{abstract}


\section{Introduction}

Digital Twins (DT) are dynamic virtual representations of physical systems that maintain a live connection with their real-world counterparts and respond to real-time changes, which supports proactive decision-making across domains such as manufacturing, healthcare, and transportation. A DT couples a physical reality, a virtual representation, and the interconnections that exchange information between them~\cite{VANDERHORN2021113524, kritzinger2018digital, jones2020characterising, fuller2020digital}. In transportation, DTs replicate elements such as pavement condition, vehicle dynamics, pedestrian behavior, and signal status from cameras, LiDARs, and traffic detectors, supporting data-driven analysis and operations.

Transportation DTs rely on high-quality data and representations, raising two challenges for real-world implementation. The first is a lack of accurate, real-time sensing of roadway infrastructure. The second is a dependence on centralized architectures that raise concerns over privacy, communication cost, and scalability.

\begin{figure}[h]
    \centering
    \includegraphics[width=0.75\textwidth]{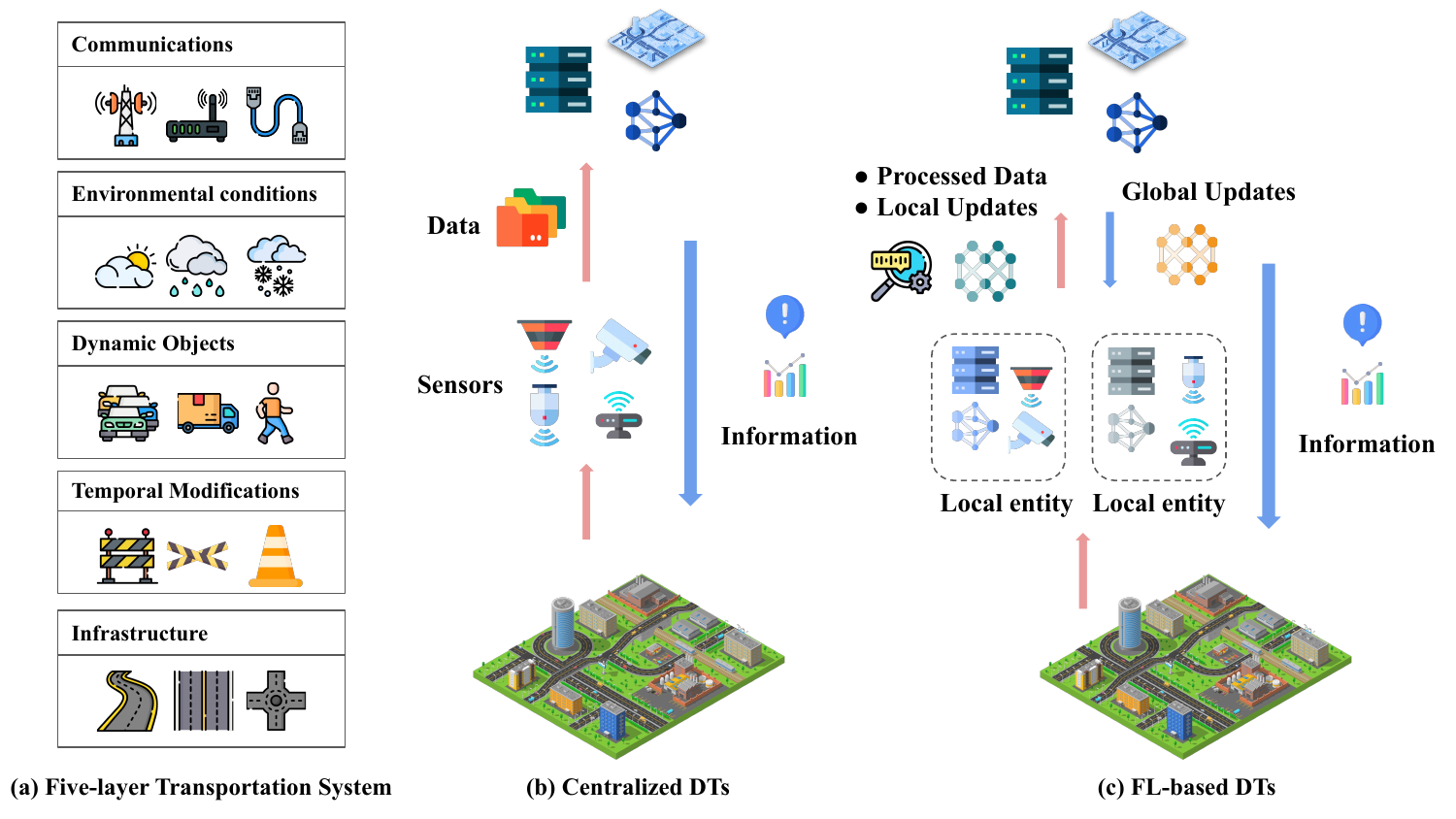}
    \caption{Transportation systems and digital twins. (a) Five-layer transportation model~\cite{scholtes20216}, from the road network up through dynamic objects and digital information. (b) Centralized DTs collect all data at a central server. (c) Federated DTs keep data on local entities and exchange only model updates with the server.}
    \label{fig:overview}
\end{figure}

Transportation DTs can be described with the five-layer model in Figure~\ref{fig:overview}(a). Recent work advanced the dynamic entities of the upper layers, including vehicles, pedestrians, signals, and communications~\cite{wang2021digital,hui2022collaboration,dong2023mixed,tan2023digital,liao2024digital,wang2024smart, cai2023,zelenbaba2022wireless, liu2022application, cazzella2024multi, WAGNER2023, dasgupta_transportation_2023, fu2024digital}. In contrast, infrastructure layers remain underexplored, relying on static maps~\cite{dasgupta_transportation_2023,wang2024smart,wang2021digital,WAGNER2023} or costly, labor-intensive LIDAR surveys~\cite{pan2024scan,jiang2022building,zhang20193d,davletshina2024automating}.

The second challenge is architectural. Centralized DTs that pool data from various sources raise concerns over privacy, communication cost, and scalability, as shown in Figure~\ref{fig:overview}(b). Federated Learning (FL) offers a decentralized alternative in which entities collaboratively train a model while keeping their data local~\cite{kairouz2021advances}, improving privacy and reducing communication cost~\cite{zhang2024survey}. As shown in Figure~\ref{fig:overview}(c), local entities such as vehicles, edge units, or roadside units train on their own data and transmit only model updates to a central server that aggregates them into a global model.

To address these challenges, we propose Geo-ORBIT (Geometrical Operational Roadway Blueprint with Integrated Twin), a decentralized, privacy-aware traffic DT framework that synchronizes real-world roadway geometry and vehicle trajectories in simulated environments. The framework utilizes a meta-learning approach for sensing roadway infrastructure, enabling local entities to contribute contextual knowledge about roadway geometry using camera data. An FL-based optimization strategy is proposed to train a global meta-learner on a central server by aggregating parameters from local entities, without transmitting private data. This global meta-learner deploys to unseen locations without re-training, which improves scalability. The central server replicates roadway and traffic conditions in a simulated environment with SUMO and CARLA, building a DT for continuous traffic monitoring, analysis, and decision-making. Our research contributions are as follows:

\begin{itemize}
    \item We develop a label-free lane detection method that infers lane geometry from roadside-camera trajectories, and show the detected geometry is approximately an order of magnitude closer to human annotations than the public map that supervises it.
    \item We introduce an annotation-based evaluation that scores detection in meters and separates the detection error from the reference-map error at each site, showing the residual error comes mostly from the map.
    \item We propose a federated meta-learning strategy that matches per-camera specialists at trained sites and transfers more reliably to held-out sites than a per-camera model, while reducing per-camera communication by more than 99 percent relative to centralizing trajectories.
    \item We demonstrate a closed-loop DT update in which a detected lane change is mapped into a running CARLA-SUMO simulation and its effect quantified, extending the twin beyond a static import.
\end{itemize}

\section{Related Work}

\subsection{Infrastructure Sensing in Digital Twins}

DTs have been developed for diverse transportation areas~\cite{wang2021digital,hui2022collaboration,dong2023mixed,tan2023digital,liao2024digital,wang2024smart, cai2023,zelenbaba2022wireless, liu2022application, cazzella2024multi, WAGNER2023, dasgupta_transportation_2023, fu2024digital,keshu2025}, and vehicle DTs have dominated due to mature onboard sensing and communication capabilities. Example applications include non-signalized intersection scheduling \cite{wang2021digital}, sand-table cooperative platooning \cite{dong2023mixed}, and collision warning fusion \cite{wang2024smart}. The infrastructure part, however, has been developed using either static maps~\cite{dasgupta_transportation_2023, wang2024smart, wang2021digital,WAGNER2023} or non-real-time sensing technologies~\cite{pan2024scan,jiang2022building,zhang20193d,davletshina2024automating}. While camera-based solutions are cost-effective~\cite{qiu2024intelligent,ren2014lane}, prior works haven't closed the loop between the physical and the digital space.

\subsection{Federated Learning and Digital Twins}

FL has been widely used in various domains, including mobile computing, industrial engineering, and health care~\cite{li2020review}. In transportation, FL is used in traffic flow prediction, traffic object detection, and vehicular edge computing~\cite{zhang2024survey}. These studies have suggested FL's advantages in preserving privacy, enhancing system scalability, promoting data diversity, and improving network efficiency.

FL-based DTs aggregate model information from local entities instead of raw data, which preserves privacy and reduces communication. Recent work extends this to networked digital twin systems, for example, multi-modal federated learning across satellite-ground twin networks~\cite{gong2025tmc}, sharing our aim of coordinating distributed twins without centralizing raw data.

Most studies focused on proposing novel FL techniques while developing FL-based DTs. \citet{sun2021} used a reinforcement learning-based FL strategy to optimize the global aggregation frequency in industrial IoTs, with DTs mapping device state and energy consumption for the optimization. \citet{mu2023} combined DT and FL for target detection through a federated transfer learning framework and a blockchain-based scheme that screens malicious entities before aggregation. \citet{tang2024} predicted vehicle collisions with a semi-asynchronous FL framework that adjusts client participation according to network conditions, vehicle resources, and mobility. \citet{khan2024} integrated DT and FL in vehicular networks with local models trained at onboard and roadside units and aggregated centrally, and discussed edge caching, resource management, and communication challenges.

In summary, existing FL-based DTs focus on the FL technique while modeling the environment with low-fidelity numerical simulation, which limits their applicability in complex real-world scenarios.

\section{The Framework for Lane Detection Algorithm with Federated Learning Integration into Digital Twin}

\begin{figure}[h]
    \centering
    \includegraphics[width=0.9\textwidth]{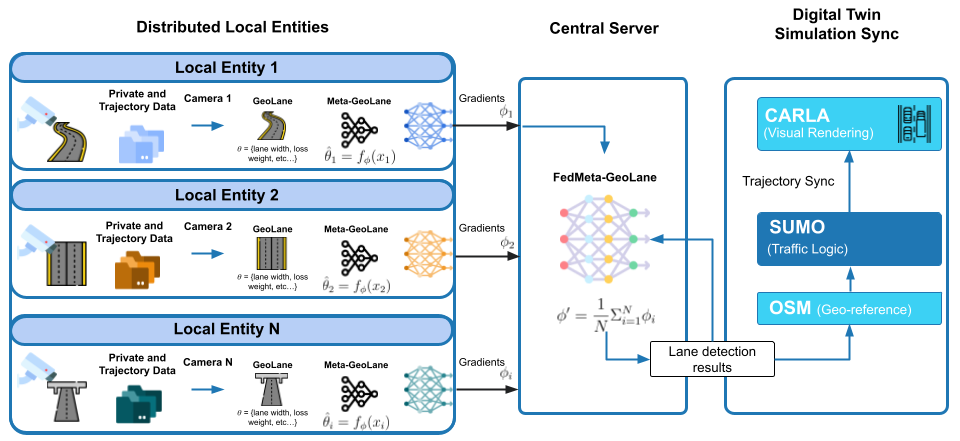}
    \caption{Architecture of the federated meta-learning framework. The framework detects roadway geometry at local entities with local GeoLane models. The central server collects parameters from local entities with federated learning. The DT synchronizes road geometry and trajectories in a simulated environment.}
    \label{fig:architecture}
\end{figure}

Geo-ORBIT aims to enhance real-time traffic management and operations by dynamically sensing, modeling, and synchronizing roadway geometry and vehicle behavior within a DT environment. As illustrated in Figure~\ref{fig:architecture}, the framework is designed to operate across multiple roadside sensing units, each of which locally processes vehicle and roadway observations to infer lane-level geometry. A meta-learning approach is employed to adapt to diverse scene contexts, while federated learning enables collaborative model optimization across locations without sharing raw data. The resulting lane geometries and trajectory information are synchronized with simulation environments to support dynamic digital twin updates for traffic management and analysis.

\subsection{Federated Meta-Learning for Lane Detection Framework}

To enable scalable and privacy-preserving deployment across diverse roadside environments, we propose a federated meta-learning framework that adapts the detection process to local scene characteristics without centralizing sensor data.

\subsubsection{Task Definition and Motivation}
We define each camera entity as a task $\mathcal{T}_i$, characterized by unique camera perspectives, road geometries, traffic dynamics, and environmental conditions. Lane detection models with fixed parameters, trained centrally, degrade across such heterogeneous contexts. We therefore train a meta-model to infer detection parameters $\theta_i$ for each task from high-level contextual features $x_i$. Unlike gradient-based meta-learning such as MAML~\cite{finn2017model}, which requires end-to-end differentiability, our black-box formulation \cite{hu2023} operates entirely at the parameter level and is compatible with the non-differentiable detection pipeline.

\subsubsection{Meta-Learner Design}
The meta-learner $f(\phi)$ is implemented as a two-layer multi-layer perceptron (MLP) with shared hidden representation and multiple parameter-specific output heads. The network takes a vector of scene-specific features as input (e.g., trajectory speed, time of day) and outputs a dictionary of detection parameters tailored to the current task $\mathcal{T}_i$.

Each head maps the shared latent representation to a scalar output using a final linear layer and appropriate activation (e.g., sigmoid, ReLU). For example, the predicted smoothing factor is mapped from $[0, 1]$ to the desired numeric range (e.g., $1-20$) via scaling. The architecture is designed to allow interpretable, bounded, and task-specific parameter estimation directly from scene features, without backpropagating through the lane detection pipeline.

Each stage of the detection pipeline in the following section is controlled by these parameters $\theta_i$, and the meta-learner is optimized using the geometric similarity metrics of Section \RN{4}.

\subsubsection{Federated Optimization Strategy}
To preserve data privacy and enable decentralized learning, we adopt a federated optimization protocol for training the meta-learner. The global meta-learner is trained by federated averaging of the client updates, so that one shared model captures the scene-to-parameter mapping across all sites while the raw data stays on each camera. At each federated round:
\begin{itemize}
    \item A subset of client nodes (roadside units) receive the current global meta-learner parameters $\phi$.
    \item Each client extracts local features $x_i$, generates pipeline parameters $\theta_i = f_\phi (x_i)$, and runs the detection pipeline.
    \item The client computes the local task loss $\mathcal{L}_i$ and a parameter update with respect to $\phi$, without backpropagating through the detection model.
    \item The parameter updates are communicated to a central server, where they are aggregated to update the global meta-learner.
\end{itemize}
This design ensures that no raw trajectory data or image content is ever transmitted, aligning with the privacy constraints inherent to distributed sensing in transportation systems.

Once trained, the meta-learner deploys to new or existing cameras. Given a novel context $x_j$ it outputs detection parameters $\theta_j = f_\phi(x_j)$ to configure the local pipeline, adapting to the scene with no retraining or fine-tuning.

\subsection{Knowledge-Based Lane Detection Algorithm}

Each local client reconstructs lane geometry from vehicle trajectories in roadside camera feeds, without high-definition maps or annotated labels, combining geometric reasoning with weak supervision to produce structured lane models for simulation and evaluation.

\begin{figure}[h]
    \centering
    \includegraphics[width=0.9\textwidth]{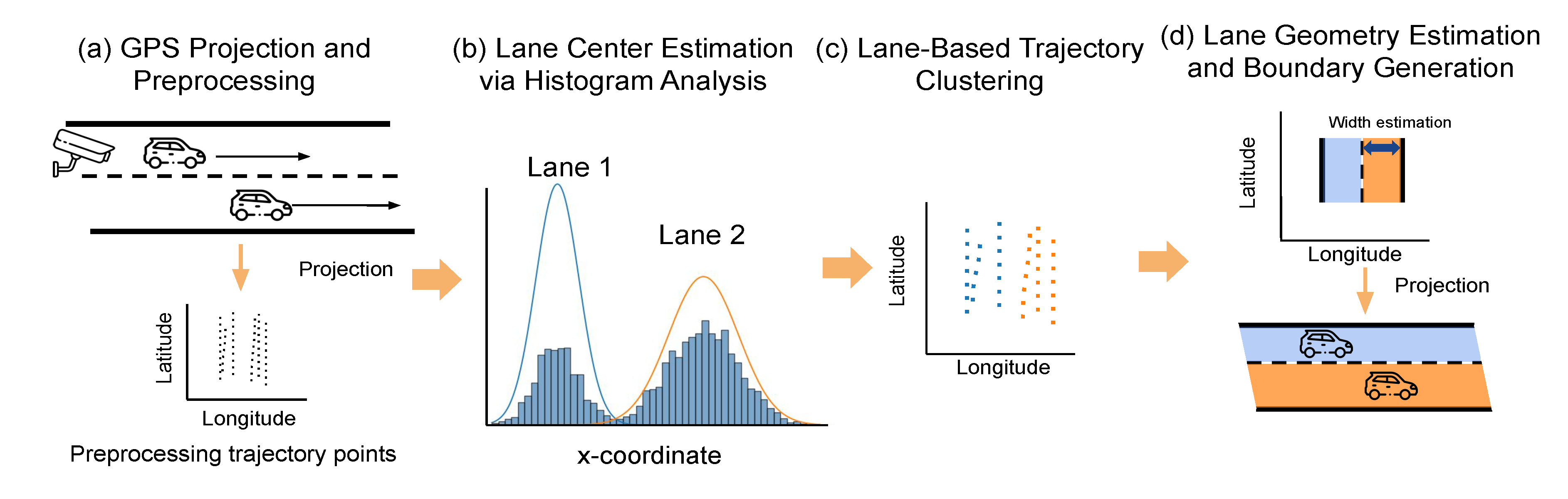}
    \caption{Overview of Knowledge-Based Lane Detection Algorithm. (a) Video detection and trajectory projection to GPS coordinates. (b) Lane center estimation using histogram analysis. (c) Lane-based trajectory clustering with KMeans. (d) Lane geometry estimation and boundary generation.}
    \label{fig:lane_detection}
\end{figure}

Figure~\ref{fig:lane_detection} illustrates the lane detection flow. Vehicles are first detected with the YOLOv11 model~\cite{Jocher_Ultralytics_YOLO_2023} over an observation period (e.g., 60 seconds) and their trajectories stored locally.

The knowledge-based lane detection proceeds through four primary stages shown in Figure~\ref{fig:lane_detection}: (a) trajectory projection to global GPS coordinates, (b) lane count estimation via histogram analysis, (c) trajectory clustering using KMeans, and (d) lane geometry estimation via spline fitting and boundary modeling.

New trajectory data are incorporated iteratively to refine lane center positions and geometries. Detected lanes are then validated against SUMO geometry by aligning the two and rectifying discrepancies, and SUMO data feed back into detection to improve accuracy.

Raw detections are projected from pixel space to GPS coordinates using homography matrices fitted from known image-to-GPS correspondences. Within each lane group, trajectories are filtered for missing data and aggregated into per-object mean positions and heading directions.

To estimate the number of lanes in each segment, the system performs histogram-based peak detection on the lateral offsets of vehicle trajectories. For each road segment, we determine the longitudinal road axis from per-vehicle displacement vectors and project each trajectory's mean position onto its perpendicular. This ensures that the coordinate entering the histogram represents the true across-road offset in meters rather than an arbitrary image or GPS axis, preventing adjacent lanes from merging on roads running diagonally across the camera frame. Let $X = \{x_i\}$ denote the set of across-road offsets within a segment. A histogram is constructed over $X$, clipped to the 1st and 99th percentiles to mitigate outlier tracks, smoothed with a Gaussian filter, and processed via peak detection.

The histogram is normalized by its maximum prior to peak finding, ensuring the learned peak prominence $\theta_{\text{prominence}}$ acts as a scale-free fraction of the local peak height rather than an absolute count that requires per-site retuning. This relative prominence formulation transfers reliably across both sparse and dense traffic conditions, enabling zero-shot deployment at held-out locations. Additionally, the smoothing bandwidth is bounded below by the histogram noise floor to prevent spurious local peaks. The resulting peaks $\{p_1, p_2, \ldots, p_k\}$ serve as preliminary lane center estimates, guiding subsequent clustering and spline fitting stages.

Based on the estimated number of lanes $k$, KMeans clustering is applied to the set of across-road offsets to assign each trajectory to a lane:

\begin{equation}
    \hat{y} = \text{KMeans}(X, \text{clusters} = k).
\end{equation}

Each vehicle is assigned to its closest lane center. A detected lane supported by fewer than $\theta_{\text{evidence}}$ trajectories is then discarded, where the minimum per-lane support $\theta_{\text{evidence}}$ is itself predicted by the meta-learner. Discarding low-support clusters, which are disproportionately spurious, raises precision relative to the ungated fixed configuration at some cost to the recall of genuinely sparse lanes, and the meta-learner sets $\theta_{\text{evidence}}$ together with the prominence to balance the two.

For each detected lane, the corresponding trajectory points are sorted by their longitudinal GPS coordinate $y_{\text{gps}}$. A univariate spline $x = f(y)$ is then fitted to the data:
\begin{equation}
    x(y) = \text{Spline}(y; s),
\end{equation}
where $s$ is a smoothing parameter governed by $\theta_{\text{smoothing}}$ and floored at the residual of a straight-line fit so that the spline does not fit the trajectory noise.

The spline is evaluated at uniformly spaced intervals to form a centerline curve. To estimate lane width, the lateral spread of the assigned trajectory points is measured:
\begin{equation}
    w = 2 \cdot \sigma_x,
\end{equation}
where $\sigma_x$ is the standard deviation of across-road offsets, representing approximately $95\%$ of the lane's lateral extent assuming a Gaussian distribution.

Boundary lines are generated by computing the unit normals $(n_x, n_y)$ along the spline curve and offsetting the centerline:
\begin{equation}
    \text{Left}(y) = (x(y), y) + \frac{w}{2}\cdot \vec{n}, \quad \text{Right}(y) = (x(y), y) - \frac{w}{2}\cdot \vec{n},
\end{equation}
where $\vec{n} = (-\frac{dy}{\sqrt{dx^2+dy^2}}, \frac{dx}{\sqrt{dx^2+dy^2}})$ is the normalized perpendicular vector.

\subsection{Digital Twin Environment Setup}

We construct a detailed DT environment by integrating Simulation of Urban MObility (SUMO) and the CARLA autonomous driving simulator, which together simulate large-scale traffic dynamics and perception-level sensor data. The base road geometry remains static during simulation while trajectories synchronize with real-world data streams. Section \RN{5} demonstrates one closed-loop exception, in which a detected lane change is applied to the running simulation without re-importing the network.

The road network is derived from OSM, converted to the SUMO network format, and populated with real vehicle trajectories. It is then converted to OpenDrive (XODR) format, which preserves geometric and semantic consistency between the traffic simulation and the 3D perception environment, and imported into CARLA for photorealistic sensor simulation including RGB camera streams.

Observed trajectories are recorded in pixel coordinates, transformed to GPS space using homography calibration, and converted to SUMO's local coordinate system with the sumolib network model. A dedicated module then maps each trajectory to the nearest valid edge in the SUMO network, generates route files from the empirical motion sequences, and injects them into the running simulation, where vehicles are repositioned at each time step to match their real-world counterparts.

\section{Lane Geometries Quantifications}
To evaluate the quality of the detected lane geometries produced by our system, we introduce a set of loss functions used in training and a set of evaluation metrics used for reporting. The training losses quantify alignment with map-based references and consistency with the meta-parameters. The reported evaluation metrics, described in Section \RN{5}, are independent of any learned weighting and are expressed in meters, so that the reported quantities are comparable across methods. We separately assess downstream trajectory alignment within a DT environment and communication efficiency in an FL context.

\subsection{Geometric and Structural Alignment Metrics}

We first assess the consistency of the detected lane shapes with reference map geometries from OSM using the Fr\'echet distance, which evaluates global curve similarity. Let $P(t)$ and $S(t)$ be continuous parameterizations of the reference and detected centerlines, respectively. The consistency loss is defined as:
\begin{equation}
    \mathcal{L}_{\mathrm{consistency}} = d(S,P)
    = \inf_{\alpha,\beta} \max_{t \in [0,1]} \bigl\lVert
        P(\alpha(t)) - S\bigl(\beta(t)\bigr)
    \bigr\rVert_2,
\end{equation}
where $\alpha$ and $\beta$ are continuous, non-decreasing reparameterizations. This formulation captures the overall shape discrepancy between two lanes.

To measure local geometric agreement, we compute the difference in lane widths between the detected lanes and their corresponding references. The geometry loss is expressed as:
\begin{equation}
    \mathcal{L}_{\mathrm{geometry}} = \sum_M \bigl(\bigl\lVert \bm{s}_{m_w} - \bm{c}_{m_w}\bigr\rVert_2^2\bigr),
\end{equation}
where $\bm{s}_{m_w}$ and $\bm{c}_{m_w}$ denote the detected and reference widths of matched lane segments. The reference width of each lane is read from the corresponding lane in the SUMO road network rather than assumed to be a fixed constant, so the comparison reflects the per-lane widths the network actually encodes.

We further include a centerline term with a triplet structure. It encourages each detected lane center $\bm{s}_m$ to lie closer to its matched reference center $\bm{c}_m$ than to the nearest other reference center $\bm{c}'_m$:
\begin{align}
    \mathcal{L}_{\mathrm{center}} = \sum_{\mathcal{E}}
    \max \Bigl\{ & \bigl\lVert c_m - s_m \bigr\rVert_2^2 \nonumber \\
                 & - \bigl\lVert c_m - c'_m \bigr\rVert_2^2, 0 \Bigr\},
\end{align}
where the negative $\bm{c}'_m$ is the nearest reference lane other than the matched one. The term is therefore a distance-based hard-negative margin computed directly on lane coordinates rather than on a learned embedding.

To enforce correct lane grouping, we define a lane number loss based on the absolute difference between the detected and reference lane counts:
\begin{equation}
    \mathcal{L}_{\text{lane\_num}} = |N_{\mathrm{det}} - N_{\mathrm{ref}}|.
\end{equation}

\subsection{Composite Objective Function}
The overall training objective combines all previously defined loss terms into a single expression, weighted by hyperparameters $\lambda_i$:
\begin{equation}
    \mathcal{L}_{\text{total}} = \lambda_1 \mathcal{L}_{\text{consistency}} + \lambda_2 \mathcal{L}_{\text{geometry}} + \lambda_3 \mathcal{L}_{\text{center}} + \lambda_4 \mathcal{L}_{\text{lane\_num}}.
\end{equation}
In the corrected evaluation protocol, the weights $\lambda_i$ are held fixed and equal across all methods, so no reported quantity depends on a learned weighting and the totals are comparable across the baseline, the per-camera meta model, and the federated model. The lane number term $\mathcal{L}_{\text{lane\_num}}$ corresponds to the exact-match and mean-absolute lane-count error reported in Section \RN{5}, and the width term $\mathcal{L}_{\text{geometry}}$ corresponds to the geometric error in meters reported there, so the training losses and the reported metrics use the same quantities under fixed weights.

\subsection{Meta-Learning Parameter Alignment}
In our federated meta-learning framework, each client predicts a set of geometric parameters $\hat{\theta}$. These are compared against reference parameters $\theta^*$, derived from weak supervision from SUMO and OSM. We define a parameter alignment loss using the mean squared error across all parameters in the set $\mathcal{P}$:
\begin{equation}
    \mathcal{L}_{\text{param}} = \sum_{p \in \mathcal{P}} \lVert \hat{\theta}_p - \theta_p^* \rVert^2_2.
\end{equation}
This alignment loss is not backpropagated through the downstream lane detection pipeline. Rather, it is used independently to supervise the meta-learner that generates candidate parameters for the black-box lane detection model. As such, $\mathcal{L}_{\text{param}}$ is optimized during a separate meta-learning phase and is not included in the composite loss used to train or evaluate lane detection accuracy.

\subsection{Communication Cost}
Finally, we evaluate the communication overhead inherent in our federated meta-learning approach. For each round of communication between clients and the global server, we measure the total number of bytes transmitted per camera, counting both the upload of the local update and the download of the aggregated global parameters. We report this payload in bytes per camera per round and, for the fleet, as the total bytes exchanged per round, and we compare it against the volume that centralizing raw trajectories would require. Communication and computation efficiency in satellite-ground integrated networks has been improved through multi-task offloading \cite{gong2025twc}, and our design reduces communication structurally by exchanging only the low-dimensional parameter vector instead of raw trajectories.

\section{Experiments and Results}

\subsection{Data and Evaluation Protocol}
The Wisconsin Department of Transportation operates over 400 roadside cameras across the state and streams their footage through the 511 Wisconsin service\footnote{\url{https://511wi.gov/}}. We evaluate the framework on twelve sites along the US 12/18 and I-43 corridors in Madison, Wisconsin, which span a range of highway and arterial geometries. For each site, we collect vehicle trajectories over a fixed observation window. Table~\ref{tab:sites} summarizes the twelve sites, which span six to twelve annotated lanes, two to four lane groups, and a wide range of traffic volume, so the held-out split tests scenes that differ from the training sites in both geometry and demand. Two remaining sites are treated as calibration case studies as described below.

\begin{table}[!h]
    \centering
    \caption{Characteristics of the twelve study sites. Lane groups and lanes are counted from the human annotations, and trajectories are the unique tracked vehicles in the 60 s observation window.}
    \label{tab:sites}
    \begin{tabularx}{0.85\textwidth}{Xcccc}
    \toprule
    \textbf{Site} & \textbf{Corridor} & \textbf{Groups} & \textbf{Lanes} & \textbf{Traj.} \\
    \midrule
    \multicolumn{5}{l}{\textit{Training}} \\
    Todd & US 12/18 & 2 & 10 & 326 \\
    Monona & US 12/18 & 2 & 10 & 330 \\
    Yahara & US 12/18 & 2 & 8 & 534 \\
    CountyAB & US 12/18 & 2 & 6 & 67 \\
    Mineral & US 12/18 & 3 & 9 & 196 \\
    University & US 12/18 & 3 & 7 & 160 \\
    \midrule
    \multicolumn{5}{l}{\textit{Held-out}} \\
    Park & US 12/18 & 2 & 11 & 284 \\
    Greenway & US 12/18 & 4 & 9 & 260 \\
    Keefe & I-43 & 3 & 8 & 281 \\
    Walnut & I-43 & 3 & 10 & 387 \\
    \midrule
    \multicolumn{5}{l}{\textit{Calibration case studies}} \\
    Stoughton & US 12/18 & 3 & 12 & 297 \\
    Whitney & US 12/18 & 2 & 6 & 351 \\
    \bottomrule
    \end{tabularx}
\end{table}

Each client, representing a distinct camera location, provides data characterized by its unique road geometry and vehicle behavior patterns. The data for each client include detected trajectory points from processed video frames and an OSM-derived reference for lane centerlines that serves as weak supervision during training.

We report every result as a mean over three training seeds, except the fixed baseline, which is deterministic and reported as a single value. Seed variation is small throughout, with the geometry-term standard deviations below 0.5 m, and the F1 standard deviations are given inline where a comparison turns on them. F1 is computed per seed and then averaged, so it is not the harmonic mean of the tabulated precision and recall. Geometric quantities are expressed in meters and computed identically for every method, and no reported metric contains any learned weighting term. This corrects a comparability issue in the originally submitted loss table, where each model's totals were scaled by that model's own learned loss weights while the baseline used unnormalized weights, so the totals were not directly comparable across methods. The primary reference for evaluation is a set of human lane annotations at the twelve sites, used for evaluation only and never for training or parameter selection. Detected lanes and annotations are projected through the same site homography, so calibration error cancels in the comparison. The site admission rule that determines this split is detailed in Section~\RN{5}-C.

\subsection{Lane Detection Accuracy}

We evaluate lane detection accuracy against the human annotations, which are independent of the OSM reference used in training. Table~\ref{tab:gt_comparison} reports precision, recall, and F1 at a 5 m match threshold for six methods on the seen and held-out sites. Detection and annotations are matched one-to-one by a Hungarian assignment, and the threshold sets the distance at which a detected lane counts as matching an annotated one. The threshold is set near a lane width to tolerate the residual homography offset, and the one-to-one assignment prevents a detection from being credited against more than one annotated lane, so a detection near two adjacent annotations is matched to at most one of them.

\begin{table}[!ht]
    \centering
    \caption{Lane-level detection against human annotations. The GeoLane-based models yield anchored geometry with widths and boundaries, while the external baselines produce centerlines only. Learned models are averaged over three training seeds.}
    \label{tab:gt_comparison}
    \begin{tabularx}{0.9\textwidth}{Xcccccc}
    \toprule
    & \multicolumn{3}{c}{\textbf{Seen}} & \multicolumn{3}{c}{\textbf{Unseen}} \\
    \cmidrule(lr){2-4} \cmidrule(lr){5-7}
    \textbf{Method} & P & R & F1 $\uparrow$ & P & R & F1 $\uparrow$ \\
    \midrule
    OSM reference        & 0.65 & 0.60 & 0.63 & 0.28 & 0.34 & 0.31 \\
    Qiu et al.~\cite{qiu2024intelligent} & 0.63 & 0.44 & 0.52 & 0.38 & 0.16 & 0.22 \\
    Ren~\cite{ren2014lane}          & 0.70 & 0.74 & 0.72 & 0.68 & 0.68 & 0.68 \\
    GeoLane (baseline)   & 0.83 & 0.60 & 0.698 & 0.81 & 0.55 & 0.656 \\
    Meta-GeoLane         & 0.89 & 0.67 & 0.760 & 0.90 & 0.49 & 0.636 \\
    FedMeta-GeoLane      & 0.88 & 0.67 & 0.757 & 0.91 & 0.53 & 0.666 \\
    \bottomrule
    \end{tabularx}
\end{table}

\textbf{Accuracy against the map.} The detected geometry is closer to the annotations than the OSM reference that supervises training. Across the ten admitted sites, the mean detection error is 2.7 m, against a mean OSM-to-annotation offset near 26 m, as decomposed in Section~\RN{5}-C. The system therefore produces lane geometry roughly an order of magnitude more accurate than the public map it learns from, without any human label in the loop. When the reference itself is scored as a detector under the same protocol, it reaches an F1 of 0.63 on the training sites and only 0.31 at held-out sites, which quantifies the staleness of the map, and every method except the Qiu ROI baseline exceeds it at held-out locations.

\textbf{Adaptation against the fixed configuration.} At the training sites, the scene-adaptive models improve on the fixed configuration on every axis that traffic evidence supports. Lane-level F1 at the 5 m threshold reaches 0.760 for the per-camera meta model, and 0.757 for the federated model against 0.698 for the fixed baseline, and the gain comes from recall, 67 and 67 against 60 percent of annotated lanes. The largest effect is on lane counting. Table~\ref{tab:geometry_decomp} reports the count error together with the consistency, center, and width terms, which are close across methods and reflect the reference floor of Section~\RN{5}-C. The fixed configuration under-counts at particular sites, with count errors of six lanes at Todd and four at Monona, and the adaptive models reduce the Monona count error from four to about one, while Todd remains bounded by trajectory coverage for all methods. The mean absolute count error falls from 2.67 to 2.00 at the training sites. At every site, the error is dominated by under-counting, lanes that carry too little trajectory evidence to be detected, not by over-detection, so the count metric is a facet of the recall limitation reported in Section~\RN{5}-F. At held-out sites, the lower count error of the fixed baseline is an artifact of cancellation: missed lanes offset admitted spurious lanes due to lower precision (0.81 vs. 0.91 for FedMeta). By contrast, FedMeta prunes low-support noise via $\theta_{\text{evidence}}$ and adaptive prominence, tracking true geometry. In terms of recall (0.53 vs. 0.55), FedMeta matches the baseline within seed noise while delivering substantially higher precision.

\begin{table}[!ht]
    \centering
    \caption{Geometry decomposition against human annotations. Consistency is the Fr\'echet distance, Center the matched centerline offset, Width the lane-width error, and Count the mean absolute lane-count error, reported as mean (SD) over three seeds with the deterministic baseline as a single value. Lower is better.}
    \label{tab:geometry_decomp}
    \begin{tabularx}{\textwidth}{Xcccccccc}
    \toprule
    & \multicolumn{4}{c}{\textbf{Seen}} & \multicolumn{4}{c}{\textbf{Unseen}} \\
    \cmidrule(lr){2-5} \cmidrule(lr){6-9}
    \textbf{Method} & Consist & Center & Width & Count $\downarrow$ & Consist & Center & Width & Count $\downarrow$ \\
    \midrule
    GeoLane (baseline) & 5.02 & 1.91 & 2.37 & 2.67 & 4.76 & 2.21 & 2.07 & \textbf{3.00} \\
    Meta-GeoLane & 5.48 & 1.96 & 2.21 & 2.06 (0.21) & 4.94 & 2.18 & 1.92 & 4.33 (0.24) \\
    FedMeta-GeoLane & 5.49 & 2.03 & 2.19 & \textbf{2.00} (0.14) & 4.95 & 2.24 & 1.91 & 4.00 (0.20) \\
    \bottomrule
    \end{tabularx}
\end{table}

\textbf{External baselines.} The comparison includes two published trajectory-based methods run on identical detections, so the differences measure lane learning rather than detection quality. The multiple-ROI system of Qiu et al.~\cite{qiu2024intelligent}, using the authors' implementation, is competitive on clean highway views and reaches 0.80 at one held-out interstate site, while it returns no lanes at two arterial sites whose geometry violates its ROI assumptions, for an overall F1 of 0.52 on training sites and 0.22 at held-out sites. The method of Ren et al.~\cite{ren2014lane} clusters vehicle trajectories by their lateral position to estimate lane centerlines directly from traffic, without a map or labels. In our implementation, with its clustering thresholds fixed before evaluation, it is a strong baseline on raw centerline recall, reaching an F1 of 0.72 at the training sites and 0.68 at held-out sites, but it does so at uniformly lower precision than our methods and returns centerlines only, without the lane widths, boundaries, or network anchoring that the simulation requires. Only the proposed system yields the complete anchored lane model that the digital twin needs. The Ren clustering performs well on the centerline subtask, which points to the histogram-based separation stage as a target for future improvement.

\textbf{Federated against per-camera adaptation.} The two adaptive models reach the same accuracy at trained sites but separate at held-out sites. The per-camera meta model transfers the configuration of the nearest training site, and its held-out F1 falls to 0.636, below the fixed baseline at 0.656, because the nearest donor is not always a good match for the target scene. The federated model instead deploys one global configuration and reaches 0.666, above both the per-camera transfer and the fixed baseline. The margin grows as the held-out sites become harder to calibrate. On the four held-out sites that pass the calibration admission rule it matches the baseline, 0.666 with a seed standard deviation of 0.018 against 0.656, within that variation, and as a robustness check outside the admitted comparison, on the full set of six that also includes the two excluded calibration case studies, it reaches 0.527 with a seed standard deviation of 0.013 against 0.495, a margin the seed variation does not cover. It succeeds at the hardest calibration site, where the fixed configuration detects no lane and the federated model reaches an F1 of 0.29. The federated model therefore transfers more reliably than either a single fixed configuration or a per-camera model, with one global model in place of a model per camera and no local training before deployment at a new site.

\begin{figure}[h]
    \centering
    \includegraphics[width=0.9\textwidth]{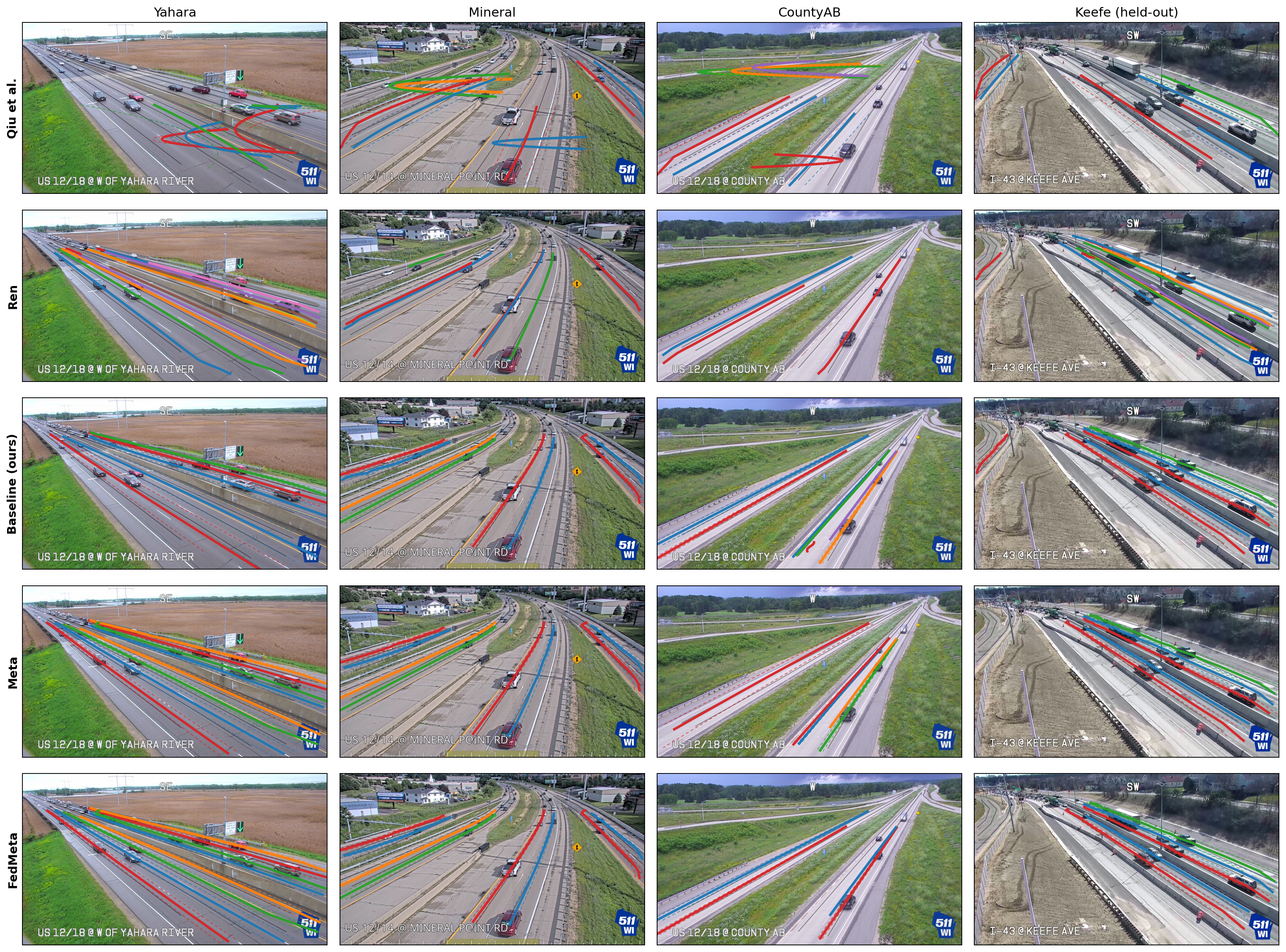}
    \caption{Qualitative comparison across four sites, including the held-out Keefe site. Detected lanes are organized by lane group and color-coded from left to right. Detections are drawn only within the extent of their respective lane groups. Estimated centerlines are solid bold, estimated boundaries are dashed, and human annotations are shown in faint black.}
    \label{fig:qualitative}
\end{figure}

Figure~\ref{fig:qualitative} visualizes the detected geometries at four sites for the five detection methods, the OSM reference of Table~\ref{tab:gt_comparison} being a static map that is not drawn. At Yahara the adaptive models detect the full multilane structure where the fixed configuration produces wandering centerlines, and at Mineral, CountyAB, and the held-out Keefe the three configurations of the proposed method produce clean, well-separated lanes with estimated boundaries, the federated model transferring to Keefe without local training. At these sites, the adaptive models also leave one lane group undetected, typically a sparsely travelled carriageway on the far side of the view, because those lanes carry little or no trajectory evidence within the observation window, so the gap is the traffic-driven recall limitation quantified in Section~\RN{5}-F not a parameter deficiency, and longer accumulation recovers such lanes as traffic arrives. Beyond these four sites, the sparse leftmost lane at Monona is detected by both adaptive models, which detect eight of the ten annotated lanes against six for the fixed configuration, the two remaining misses being lanes with no observed traffic in the recording, and at Todd limited trajectory coverage bounds recall for all methods, not a parameter deficiency, which we quantify below.

\subsection{Calibration and Reference Decomposition}

We turn the calibration concern into a measured analysis. We first corrected a defect in the homography fitting in which the robust estimator threshold was expressed in target units and therefore never rejected a control point. With the corrected estimator, a deterministic robust fit, and a geometric sanity gate, the calibration is stable across runs and rejects mis-picked control points. We report a leave-one-out reprojection audit that flags, for each site, the control points that most degrade the fit.

More importantly, we use the human annotations to decompose each site's apparent geometric error into a detection component and a reference component. Detection error, measured with annotations and detections passed through the same homography so that calibration cancels, has a cross-site mean of 2.7 m over the ten admitted sites, while the offset between the OSM reference and the annotations ranges from 2.6 to 65.9 m with a cross-site mean near 26 m, as summarized in Table~\ref{tab:calibration}. The residual site error is therefore dominated by the schematic OSM reference rather than by calibration or detection, which is why geometry does not separate the learned models and why the paper reports lane count, deployment stability, and communication as the discriminating axes.

The same audit yields a uniform, model-independent site admission rule. A site enters the train and test split when its annotation-based detection error under the fixed configuration is below 5 m. Two sites fail the rule, at 6.6 and 7.0 m, both attributable to ground control points picked near the horizon where the homography is least constrained, and we retain them as calibration case studies rather than in the comparison tables. Every admitted site measures between 0.2 and 4.1 m, and the rule keys on measured detection error rather than on any property of a particular model.

\begin{table}[!h]
    \centering
    \caption{Detection-versus-reference decomposition and the site admission rule.}
    \label{tab:calibration}
    \begin{tabularx}{0.85\textwidth}{Xcc}
    \toprule
    \textbf{Sites} & \textbf{Detection error (m)} & \textbf{Reference offset (m)} \\
    \midrule
    Admitted (10 sites) & 0.2--4.1 (mean 2.7) & 2.6--65.9 (mean 26) \\
    Excluded, Stoughton & 6.6 & -- \\
    Excluded, Whitney   & 7.0 & -- \\
    \bottomrule
    \end{tabularx}
\end{table}

\textbf{Evaluation caveats.} Three properties of this evaluation bound its claims. First, because the detections and the annotations pass through the same homography, the reported detection error is measured relative to that homography rather than as an absolute georeferenced accuracy, and a systematic calibration error at a site shifts both together and cancels in this metric. Second, the annotations are drawn by a single annotator in the image plane and carry no inter-annotator agreement, so the reference itself has an unquantified label uncertainty. Third, the trajectories come from a shared detection and tracking stage, so a lane the tracker never observes places a recall ceiling on every method that is a property of the sensing rather than of the lane model. The admission rule keys on the annotation-based detection error under the fixed configuration, so it remains independent of any learned model.

\subsection{Transmission Cost Analysis}

To evaluate the scalability of each approach, we compare the communication payload of the federated framework against centralizing raw trajectories, as summarized in Table~\ref{tab:bps}. The federated meta-learner exchanges only the low-dimensional parameter vector. Each camera uploads 632 B and downloads 262 B per round, about 0.9 KB per camera per round, and across the fleet a round costs 5.4 KB. Centralizing raw trajectories instead requires 0.35 to 4.3 MB per camera depending on traffic, or 12.9 MB for the fleet per round. The parameter-only exchange therefore reduces the per-round communication by more than 99 percent while transmitting no raw trajectory or image content, which preserves the privacy of the observed drivers.

\begin{table}[!h]
    \centering
    \caption{Communication payload per federated round. The federated framework exchanges parameters only, while centralization transmits raw trajectories.}
    \label{tab:bps}
    \begin{tabularx}{0.85\textwidth}{Xcc}
    \toprule
    \textbf{Approach} & \textbf{Per camera} & \textbf{Fleet} \\
    \midrule
    Centralized (raw trajectories) & 0.35--4.3 MB & 12.9 MB \\
    FedMeta (parameters) & \textbf{0.9 KB} & \textbf{5.4 KB} \\
    \midrule
    Reduction & \multicolumn{2}{c}{$>99\%$} \\
    \bottomrule
    \end{tabularx}
\end{table}

\subsection{Digital Twin Synchronization and Closed-Loop Update}
\begin{figure}[h]
    \centering
    \includegraphics[width=0.9\textwidth]{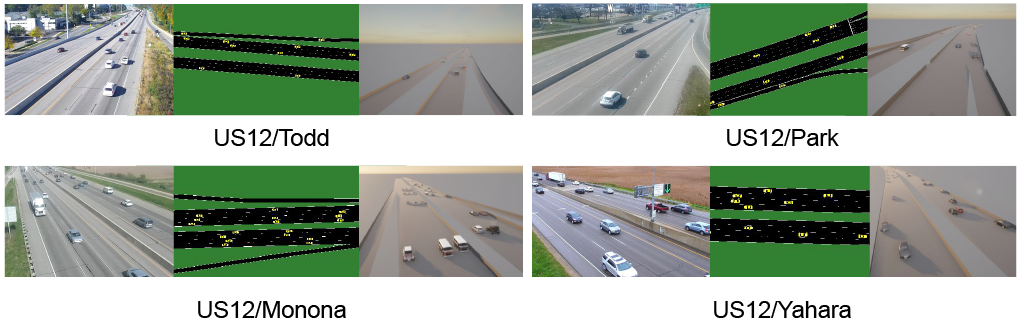}
    \caption{Digital twin synchronization with SUMO and CARLA at multiple locations employing real-time vehicle trajectory.}
    \label{fig:dt}
\end{figure}

Figure~\ref{fig:dt} illustrates the real-time synchronization between real-world vehicle trajectories and their digital representations within our integrated SUMO-CARLA DT environment. CARLA renders road segments from the perspective of roadside cameras while SUMO simulates vehicle movement, and the perception module processes either real video feeds or CARLA's rendered frames so that visually perceived behavior can be compared side by side with the simulation-grounded truth for validating perception models and digital twin consistency.

We further demonstrate one closed-loop dynamic update, which addresses the concern that the road infrastructure remains static once imported from OSM. At the Monona site, we synthesize a lane closure in the input data by removing the vehicles of one detected lane, disclosed explicitly since no physical closure occurred during the recording, while the loop that responds to it is exercised in full. The change detector compares the detected lane sets of consecutive observation windows and flags the disappeared lane, the flag carries the lane's mapping into the SUMO network through travel direction and lateral rank, which are invariant to the schematic OSM offset of Section~\RN{5}-C, and the closure is applied to the running simulation through TraCI without any re-import of the network and re-simulated under identical demand and a fixed seed. The closure raises the mean departure delay from 34.1 to 36.2 seconds, a six percent increase, and reduces the vehicles served from 2964 to 2946, a modest and entry-concentrated penalty. The demand is the osmWebWizard trip set scaled by a factor of sixty, which we state for reproducibility. Broader dynamic updates remain future work.

\subsection{Ablation and Sensitivity}

\textbf{Traffic volume.} The volume sweep bounds the method's dependence on traffic volume. Retraining with a seeded fraction of each site's trajectories over three seeds, detection F1 against the annotations changes by less than 0.02 when training uses one tenth of the recorded trajectories, and at no fraction falls more than 0.02 below the full-data value. Reliable detection therefore persists far below the recorded demand, and the binding constraint is lanes that receive no traffic at all within the observation period rather than overall traffic volume, which longer accumulation addresses.

\textbf{Baseline sensitivity.} The Ren baseline relies on hand-set thresholds, which we fixed before evaluation, so its reported row could in principle depend on those particular values. To show that it does not, we sweep the clustering threshold by plus or minus twenty percent and the lane-membership floor from two to five percent, which are the two thresholds that govern how many lanes it forms. Across this range, its F1 stays between 0.64 and 0.78 at the training sites and between 0.64 and 0.72 at held-out sites. The baseline is therefore stable to the threshold choice, and the single row we report is representative rather than a favorable pick.

\subsection{Discussions}

\subsubsection{Data-Driven Approach on Lane Detection}
The principal limitation of the method is recall. The methods detect between 49 and 67 percent of the annotated lanes, and the remaining misses are lanes with little or no trajectory evidence in the observation period, such as the two untravelled lanes at Monona, which no parameter setting detects. A global trajectory-clustering stage, as in the Ren baseline, and a learned temporal accumulation depth are natural next steps, and longer accumulation windows directly address lanes that receive no traffic within a single window.

\subsubsection{Calibration Sensitivity}
Calibration enters through the pixel-to-GPS homography, which is vulnerable to camera misalignment, imperfect manual calibration, and limited ground control points. Section~\RN{5}-C shows it is not the binding error source at admitted sites, while the two excluded sites fail on control points near the horizon. Known camera intrinsics, LiDARs, or survey-grade GNSS would reduce this further under steep curvature or elevation change.

\subsubsection{Environmental and Simulation Constraints}
The DT does not yet model physical context such as vegetation, roadside structures, and weather, which influence driver behavior and occlusion. The annotations, though independent of the training reference, were drawn by a single annotator without inter-annotator agreement, so the reference carries an unquantified label uncertainty. Quantifying that uncertainty with multiple annotators, broadening the dynamic scenarios, and extending the annotations to more corridors remain important directions.

\section{Conclusions}
This work presents Geo-ORBIT, a decentralized, privacy-aware framework for lane detection that bridges real-time perception and DT modeling. By treating each roadside camera as a local entity and learning detection parameters via a privacy-preserving meta-learner, we detect lane geometry without human labels and evaluate it against independent human annotations. The detected geometry is within approximately 3 m of the annotations at the ten admitted sites, roughly an order of magnitude closer than the public map that supervises training. The federated model matches per-camera specialists at trained sites and transfers more reliably than they do to held-out sites, where one global model improves on the fixed baseline at the hardest locations, reducing per-camera communication by more than 99\%.

The evaluation localizes the remaining error to the schematic reference and not the detection, which is why the methods separate on lane-count reliability, deployment stability, and communication cost, their geometric accuracy being bounded by the same reference map. Lane-count estimation and sparse-traffic recall remain the open challenges, and they point to a global trajectory-clustering separation stage and adaptive recall parameters as the next steps. We also plan to extend the DT beyond the single demonstrated closed-loop update, toward multi-scenario simulations enriched with environmental elements, enabling real-time traffic management and planning tools for transportation agencies.

\newpage

\bibliographystyle{unsrt}
\bibliography{reference}

\end{document}